\documentclass{esannV2}
\usepackage[latin1]{inputenc}
\usepackage{amssymb,amsmath,array}
\usepackage{booktabs}
\usepackage{graphicx}

\begin{document}
\title{Stable Diffusion Dataset Generation for Downstream Classification Tasks}

\author{Eugenio Lomurno$^1$, Matteo D'Oria$^1$ and Matteo Matteucci
%
\thanks{This paper is supported by the FAIR (Future Artificial Intelligence Research) project, funded by the NextGenerationEU program within the PNRR-PE-AI scheme (M4C2, investment 1.3, line on Artificial Intelligence).\\
\hspace*{1.5em}$^1$The authors have contributed in equal measure.}
%
\vspace{.3cm}\\
%
Department of Electronics, Information, and Bioengineering\\
Politecnico di Milano, Milan, Italy\\
\{eugenio.lomurno, matteo.doria, matteo.matteucci\}@polimi.it
}

\maketitle

\begin{abstract}
Recent advances in generative artificial intelligence have enabled the creation of high-quality synthetic data that closely mimics real-world data. This paper explores the adaptation of the Stable Diffusion 2.0 model for generating synthetic datasets, using Transfer Learning, Fine-Tuning and generation parameter optimisation techniques to improve the utility of the dataset for downstream classification tasks. We present a class-conditional version of the model that exploits a Class-Encoder and optimisation of key generation parameters. Our methodology led to synthetic datasets that, in a third of cases, produced models that outperformed those trained on real datasets. 
\end{abstract}

\section{Introduction}
\vspace{-5pt}
In recent years, the field of artificial intelligence has experienced a significant expansion, both in terms of its applications and the attention it has attracted. This surge, which shows no signs of abating, has been driven by significant advances in generative deep learning in the domains of natural language processing and computer vision. These developments have culminated in models that can synthesise data of such high quality that it is indistinguishable from real data generated by humans or from natural environments. In particular, the production of multimedia content, including images and videos, has been greatly enhanced and advanced mainly by Stable Diffusion models~\cite{rombach2022high}.
Although the application of such models is often focused on producing single samples of exceptionally high perceptual quality, an emerging trend is focused on creating synthetic datasets intended to augment or replace their real-world counterparts in downstream machine learning tasks. This change offers many opportunities, such as solving many problems related to data scarcity, the possibility of sharing and exchanging data via generators, and the improvement of data protection~\cite{lomurno2022sgde}. However, it also poses several challenges, including the reduced information content of synthetic data compared to real data, the complexity of training generative models with limited resources, and, particularly in the case of diffusion-based models, the long time required to generate a large number of synthetic samples.

In this paper, we present a study on the class-conditioned adaptation of pre-trained Stable Diffusion 2.0 model for image generation, aiming at efficiently producing synthetic datasets with high information content for a downstream classification task. The main contributions of this study can be summarised as:
\begin{itemize}
    \item We present a class-conditioned version of the Stable Diffusion 2.0 model to generate highly informative synthetic datasets from class vectors for downstream classifiers.
    \item We identify and exploit the key hyper-parameters for dataset generation, and propose an adaptation pipeline -- including transfer learning, fine-tuning and Bayesian optimisation steps -- applicable to Stable Diffusion models. 
    \item We demonstrate that our approach incrementally improves the quality of the generated datasets and reduces the per-sample generation times compared to the original model. In some cases, classifiers trained on synthetic data perform better than those trained on real data.
\end{itemize}

\section{Related Works}
\vspace{-5pt}
In recent years, there has been a notable increase in the use of generative models to augment or replace real datasets. Ravuri and Vinyals introduced the metric known as the Classification Accuracy Score (CAS) to assess the quality of synthetically generated data. This metric is designed to assess the accuracy achieved by a model trained solely on synthetic data when tested on real data not involved in the training of the generator~\cite{ravuri2019classification}. The CAS was used by Dat \textit{et al.} and Lampis \textit{et al.} to assess the effectiveness of the generation of the BigGAN model, which was trained from scratch. They also proposed several post-processing techniques to improve the utility of the generated data~\cite{dat2019classifier, lampis2023bridging}.
At the same time, other generative models have been used to generate datasets through the development of textual prompts for conditioning~\cite{sariyildiz2023fake}, to distill the information contained in real datasets into a limited number of representative samples~\cite{cazenavette2022dataset}, or to facilitate the exchange of synthetic data in federated environments as an alternative to gradients or sensitive information~\cite{lomurno2022sgde}.

\section{Method}
\vspace{-5pt}
This section presents the procedure for adapting the Stable Diffusion (SD) 2.0 model, initially pre-trained on ImageNet-1K, to generate a synthetic dataset for a classification task. Since SD 2.0 was primarily designed for text-to-image generation, the first step in facilitating class-conditioned generation is to replace the text-conditioned embedding. Consequently, the Text-Encoder is replaced by a fully-connected Class-Encoder. It has the goal to linearly map the one-hot encoded class vectors into the same dimensional space as that originally used for text conditioning in the SD 2.0 model. The proposed pipeline is then presented in four sub-steps:
\begin{enumerate}
\item \textbf{Class-Encoder Transfer Learning.} The Class-Encoder is specifically trained on the target dataset, while the rest of the architecture retains its original pre-trained weights. The training process spans 50 epochs, with a constant batch size of 64. The Adam optimiser is used with a weight decay of $4 \times 10^{-3}$, a global clip norm of 10, and a cosine annealing decay to adjust the learning rate, which is initially set to $1 \times 10^{-4}$. At the end of each epoch, a checkpoint of the Class-Encoder is stored. The primary goal of this training phase is to achieve class-conditioned generation with minimal modifications to the existing model framework.
\item \textbf{Initial Hyper-Parameters Optimisation.} The SD generation process relies heavily on two hyper-parameters: the number of Inference Steps (IS) and the Unconditioned Guidance Scale (UGS). Their tuning is crucial to obtain a good trade-off between image quality and intra-class diversity. As in the original work the goal was to maximise the single images quality with text-conditioning, using the default values -- \(IS=50\) and \(UGS=7.5\) -- would lead to sub-optimal results. For this reason, we look for their best combination, and at the same time for the best Transfer Learning epoch. To perform the tuning we use the Tree-Structured Parzen Estimator as a Bayesian optimisation strategy with Hyperband Pruning~\cite{optuna_2019}. At each tuning step, a combination of the 3 hyper-parameters is selected -- \(IS\in[5,50], UGS\in[0,7.5], Epoch\in[1,50]\) -- and a small dataset of 4000 images with uniform classes distribution is generated. This dataset is used to train a ResNet20 architecture and to compute the Classification Accuracy Score (CAS) metric~\cite{ravuri2019classification}, i.e. the Accuracy achieved by a model trained exclusively on generated data against a real test set. The tuning step is repeated for 50 iterations with the objective to maximise the CAS.
\item \textbf{Diffusion Model Fine-Tuning.} Using the optimal parameters identified at the end of the previous step, the SD model is now fine-tuned, leaving all other components unchanged during the process, including the Class-Encoder. For this step, the same training strategy is used as in step 1 except that the initial learning rate is set to $1 \times 10^{-5}$, the number of epochs is set to 10, and the batch size is set to 16. At the end of each epoch, a checkpoint of the SD model is stored.
\item \textbf{Final Hyper-Parameters Optimisation.} As in step 2, the search for the optimal adaptation epoch is carried out together with the search for the best hyper-parameters, IS and UGS. The main difference is that the number of fine-tuning checkpoints is 10, the possible values of IS are between 5 and the optimal value obtained in step 2, and the value of UGS is between 0 and twice the optimal value found previously. 
\end{enumerate}
Once the optimal configuration is identified, synthetic datasets are generated to replicate the size of the target dataset multiplied by various scaling factors ranging from 1 to 10 in integer steps. In this final step, the class distribution is kept identical to that of the real data. For each dataset, the CAS is calculated and then compared with the test accuracy of the ResNet20 model trained only on the real training set.

\begin{table}[t]
    \centering
    \setlength{\tabcolsep}{3.25pt}
    \footnotesize
    \caption{Top-1 Accuracy and Generation Time computed after each pipeline step using a synthetic dataset of 4000 images. For each dataset and metric, the best score is highlighted with \textbf{bold}.}
    \begin{tabular}{lccccccc}
    \toprule
     &  \multicolumn{4}{c}{Classification Accuracy Score (\%)} & \multicolumn{3}{c}{Generation Time (s)}\\ 
     Dataset        & After 1. &  After 2. &  After 3. & After 4. & After 1. &  After 2. or After 3. & After 4.\\ 
     \midrule
     CIFAR10        & 39.52 &  47.05 &  51.94 &  \textbf{59.30} & 30000 & \textbf{18600} & \textbf{18600}\\ 
     CIFAR100       & 15.50 &  17.48 &  17.35 &  \textbf{27.36} & 30000 & 21000 & \textbf{16800}\\ 
     PathMNIST      & 39.95 &  63.06 &  59.18 &  \textbf{78.95} & 67500 & \textbf{64800} & \textbf{64800}\\ 
     DermaMNIST     & 19.10 &  62.64 &  61.84 &  \textbf{65.43} & 5250 & 5250 & \textbf{2835}\\
     BloodMNIST     & 73.89 &  74.97 &  78.98 &  \textbf{86.20} & 9000 & 8460 & \textbf{7920}\\ 
     RetinaMNIST    & 37.00 &  40.75 &  41.74 &  \textbf{47.45} & 810 & 600 & \textbf{486}\\ 
     \bottomrule
    \end{tabular}
    \label{tab:pipelineacc}
    \vspace{-5pt}
\end{table}

\begin{figure}[t]
    \centering
    \includegraphics[width=1.\textwidth]{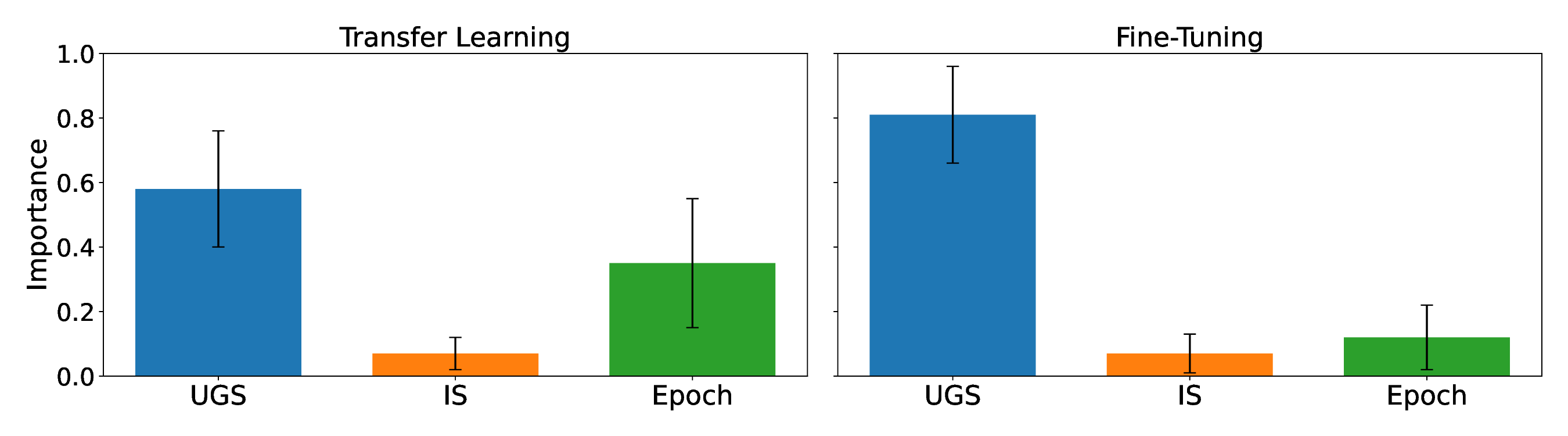}
    \vspace{-25pt}
    \caption{The hyper-parametes average importance for the two optimisation steps.}
    \label{fig:hyperparameters}
    \vspace{-5pt}
\end{figure}

\section{Experiments and Results}
\vspace{-5pt}
Experiments to assess the effectiveness of the proposed method were performed on six 32x32 RGB datasets: CIFAR10, CIFAR100, PathMNIST, DermaMNIST, BloodMNIST and RetinaMNIST. The only pre-processing is to adapt these datasets to the format expected by the SD model. Specifically, each image was resized to a resolution of 128x128 by bilinear interpolation and then normalised to the range [-1, 1]. The ResNet20 models were consistently trained over 100 epochs using an Adam optimiser with a learning rate of $1 \times 10^{-3}$, label smoothing set to 0.1, a batch size of 256, and a scheduler that reduces the rate upon reaching a plateau with a patience of 10 epochs and a reduction factor of 0.1. Early stopping was implemented with a patience of 25 epochs. No augmentation techniques were used to ensure a fair comparison between the information in the real and generated images. All experiments were performed on a single Nvidia Quadro RTX 6000.

\begin{table}
    \centering
    \setlength{\tabcolsep}{3pt}
    \caption{Top-1 Accuracy on the same real test set achieved by ResNet20 models trained on real data and its synthetic version with increasing cardinality. The best score is highlighted with \textbf{bold}, the best score from synthetic training set is highlighted with \underline{underline}.}
    \footnotesize
    \begin{tabular}{lccccccccccc}
        \toprule
        Dataset & Real &  x1 &  x2 &  x3 &  x4 &  x5 &  x6 &  x7 &  x8 & x9 & x10\\ 
        \midrule
        CIFAR10 &  83.26 &  79.09 &  81.49 &  83.67 &  83.82 &  83.67 &  84.52 &  \textbf{\underline{85.82}} &  85.47 & 85.57 & 85.79\\
        CIFAR100 &  \textbf{60.55} &  45.19 &  47.68 &  48.93 &  49.50 &  51.51 &  51.46 &  51.16 &  52.54 &  52.73 & \underline{52.93}\\
        PathMNIST &  \textbf{88.70} &  85.19 &  84.52 &  81.36 &  84.90 &  81.33 &  84.03 &  83.90 &  82.18 &  81.84 & \underline{86.45}\\
        DermaMNIST &  \textbf{75.71} &  66.98 &  67.03 &  66.93 &  68.57&  68.47&  67.38 &  \underline{69.42} &  67.38 &  68.23 & 67.73\\
        BloodMNIST &  \textbf{96.22} &  85.56 &  84.83 &  84.80 &  84.89 &  \underline{86.58} &  85.85 &  86.29 &  82.58 &  85.32 & 83.02\\
        RetinaMNIST &  47.75 &  40.50 &  43.25 &  47.00 &  \textbf{\underline{52.00}}&  47.00 &  43.00 &  51.25 &  46.25 &  49.00 & 46.50\\ 
        \bottomrule
    \end{tabular}
    \label{tab:CAS_complete_1}
    \vspace{-5pt}
\end{table}

Table~\ref{tab:pipelineacc} shows the performance achieved during each step of the adaptation pipeline by a ResNet20 model trained on 4000 synthetically generated images, evenly stratified by class.
The first evaluation, following the Transfer Learning phase (After 1), serves as a baseline for both the CAS of the classifier and the generation time of the SD model. Significant improvements are observed after the first hyper-parameter optimisation (After 2.), where CAS increases range from +1.08\% to +43.54\% and generation times decrease by up to 38\%, aided by the identification of the optimal IS value.
The Fine Tuning phase (After 3.) shows mixed results in CAS, suggesting that the parameters optimal in the previous phase may not be as effective here. However, after the final step of the pipeline (After 4.), further improvements in CAS are recorded, with increases from +10.45\% to +46.33\%, associated with reductions in generation times.

Figure~\ref{fig:hyperparameters} shows the average importance of the hyper-parameters across the two optimisation phases, using functional ANOVA to assess the importance of individual hyper-parameters and their interactions~\cite{hutter2014efficient}. While the analysis does not yield identical rankings for each phase, it consistently identifies the UGS as the most important parameter. The UGS, which is closely related to the conditioning temperature, is crucial for class-conditioned generation, and this result makes perfect sense. Contrary to our expectations, the second most important parameter is not the number of inference steps, which is typically presented as one of the most important SD parameters, but the adaptation epoch.

The achieved CAS results concerning the whole synthetic datasets generated at the end of the pipeline are reported in Table~\ref{tab:CAS_complete_1} for Top-1 Accuracy.
The synthetic datasets with the same cardinality of the training sets do not reach the CAS of the real counterpart. Nevertheless, the obtained CAS is sufficient to state that Stable Diffusion has been effectively adapted to generate a synthetic version of each target dataset.
As one scales up the size of the datasets generated, improvements can be observed. As a matter of fact, the best results are obtained from synthetic datasets with cardinality between 4 and 10 times that of the real counterpart, even with important improvements in terms of CAS.
The undoubtedly most surprising results are those for the CIFAR10 and RetinaMNIST datasets, for which models trained on synthetic datasets perform better than models trained on real data. If for the latter dataset it is possible to attribute this improvement to data scarcity, in the former case it can be generalised to the assumption that the SD model is indeed able to compete with a complete and well-structured dataset. As far as the other results are concerned, it can be observed that there is often room for improvement as the best results are close to the maximum cardinality used, thus suggesting the scalability of the approach presented.

\section{Conclusions and Future Directions}
\vspace{-5pt}
This paper explored the adaptation of the pre-trained Stable Diffusion 2.0 model using ImageNet-1K to generate synthetic datasets with high information content. The adaptation process involved a pipeline of Transfer Learning, Fine Tuning and generation parameter optimisation with the aim of improving the performance of downstream classifiers trained on the synthetic data and evaluated on real data.
Notably, in a third of cases, models trained on synthetic data outperformed those trained on real data. This highlights the potential and promising future of generative models and synthetic data, which are increasingly present in both research and industry. Future directions include extending the pipeline to include additional refinement steps, integrating techniques for filtering and post-processing generated data, and incorporating active learning strategies.

\bibliography{references}

\begin{thebibliography}{1}

\bibitem{optuna_2019}
T.~Akiba, S.~Sano, T.~Yanase, T.~Ohta, and M.~Koyama.
\newblock Optuna: A next-generation hyperparameter optimization framework.
\newblock In {\em KDD}. ACM, 2019.

\bibitem{cazenavette2022dataset}
G.~Cazenavette, T.~Wang, A.~Torralba, A.~A. Efros, and J.-Y. Zhu.
\newblock Dataset distillation by matching training trajectories.
\newblock In {\em CVPR}. IEEE/CVF, 2022.

\bibitem{dat2019classifier}
P.~T. Dat, A.~Dutt, D.~Pellerin, and G.~Qu{\'e}not.
\newblock Classifier training from a generative model.
\newblock In {\em CBMI}. IEEE, 2019.

\bibitem{hutter2014efficient}
F.~Hutter, H.~Hoos, and K.~Leyton-Brown.
\newblock An efficient approach for assessing hyperparameter importance.
\newblock In {\em ICML}. PMLR, 2014.

\bibitem{lampis2023bridging}
A.~Lampis, E.~Lomurno, and M.~Matteucci.
\newblock Bridging the gap: Enhancing the utility of synthetic data via post-processing techniques.
\newblock {\em BMVC}, 2023.

\bibitem{lomurno2022sgde}
E.~Lomurno, A.~Archetti, L.~Cazzella, S.~Samele, L.~Di~Perna, and M.~Matteucci.
\newblock Sgde: Secure generative data exchange for cross-silo federated learning.
\newblock In {\em AIPR}. ACM, 2022.

\bibitem{ravuri2019classification}
S.~Ravuri and O.~Vinyals.
\newblock Classification accuracy score for conditional generative models.
\newblock {\em NeurIPS}, 2019.

\bibitem{rombach2022high}
R.~Rombach, A.~Blattmann, D.~Lorenz, P.~Esser, and B.~Ommer.
\newblock High-resolution image synthesis with latent diffusion models.
\newblock In {\em CVPR}. IEEE/CVF, 2022.

\bibitem{sariyildiz2023fake}
M.~B. Sar{\i}y{\i}ld{\i}z, K.~Alahari, D.~Larlus, and Y.~Kalantidis.
\newblock Fake it till you make it: Learning transferable representations from synthetic imagenet clones.
\newblock In {\em CVPR}. IEEE/CVF, 2023.

\end{thebibliography}
\bibliographystyle{abbrv}
\end{document}